\pdfoutput=1

\documentclass[11pt]{article}

\usepackage{EMNLP2022}
\usepackage{flushend}

\usepackage{times}
\usepackage{latexsym}

\usepackage[T1]{fontenc}

\usepackage[utf8]{inputenc}
\usepackage{enumitem}
\usepackage{microtype}

\usepackage{inconsolata}

\usepackage{booktabs}
\usepackage{multirow}
\usepackage{amsmath}
\usepackage{microtype}
\usepackage{graphicx}
\usepackage{amsfonts}
\usepackage{tabularx}
\usepackage{color}
\usepackage{comment}
\usepackage{amsmath,amsfonts,amssymb}
\usepackage{arydshln}
\usepackage{tablefootnote}
\usepackage{xcolor}
\DeclareSymbolFont{extraup}{U}{zavm}{m}{n}

\DeclareMathSymbol{\varheart}{\mathalpha}{extraup}{86}
\title{Better Few-Shot Relation Extraction with Label Prompt Dropout}

\author{Peiyuan Zhang \and Wei Lu\\
  StatNLP Research Group\\
  Singapore University of Technology and Design \\
  \texttt{peiyuan\_zhang@sutd.edu.sg, luwei@sutd.edu.sg} \\}

\begin{document}
\maketitle
\begin{abstract}
Few-shot relation extraction aims to learn to identify the relation between two entities based on very limited training examples.
Recent efforts found that textual labels (i.e., relation names and relation descriptions) could be extremely useful for learning class representations, which will benefit the few-shot learning task.
However, what is the best way to leverage such label information in the learning process is an important research question.
Existing works largely assume such textual labels are always present during both learning and prediction.
In this work, we argue that such approaches may not always lead to optimal results.
Instead, we present a novel approach called {\em label prompt dropout}, which randomly removes label descriptions in the learning process.
Our experiments show that our approach is able to lead to improved class representations, yielding significantly better results on the few-shot relation extraction task.\footnote{Code available at \url{https://github.com/jzhang38/LPD}}
\end{abstract}

\section{Introduction}

Enabling machines to comprehend sentences and extract relations between entities has been a crucial task in Natural Language Processing (NLP). Conventional methods frame this task as a multi-class classification problem, trying to solve it through large-scale supervised training with LSTM \cite{10.1162/neco.1997.9.8.1735} or BERT \cite{devlin-etal-2019-bert} as the backbone \citep{zhou-etal-2016-attention, zhang-etal-2017-position, yamada-etal-2020-luke}. Such an approach has shown great effectiveness. However, one problem left unsolved is to identify novel relations with only a handful of training examples. Therefore, recent studies \citep{han-etal-2018-fewrel, gao-etal-2019-fewrel} introduce the task of few-shot relation extraction (FSRE) to study this data scarcity problem.

 Aligned with the success of few shot learning in Computer Vision \cite{DBLP:conf/cvpr/SungYZXTH18, DBLP:conf/iclr/SatorrasE18}, most attempts in FSRE adopt a meta learning framework \citep{DBLP:conf/icml/SantoroBBWL16, DBLP:conf/nips/VinyalsBLKW16} that randomly samples episodes with different label sets from the training data to mimic the few shot scenario in the testing phase. As a meta learning approach, prototypical network \citep{DBLP:conf/nips/SnellSZ17} aims to learn a class-agnostic metric space. A query instance is classified as the class that has the nearest prototype during inference.

\begin{figure}[t] 
		\flushleft
		\includegraphics[width=\linewidth, clip=true, trim = 0mm 20mm 10mm 20mm]{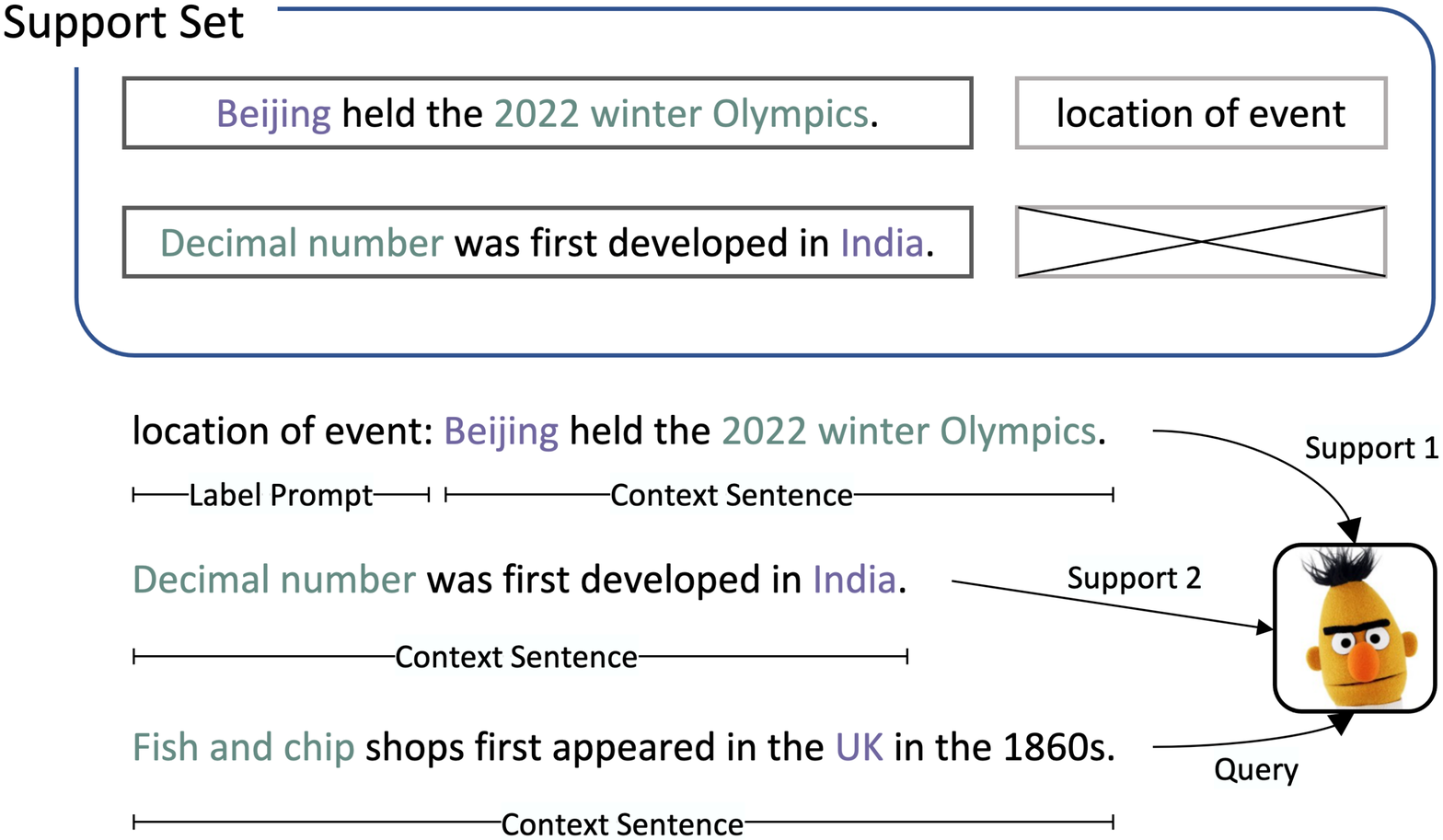}
		\caption{An example of 2-way-1-shot learning using label prompt dropout (LPD). Top: Instead of assuming textual labels are always present for support instances, LPD randomly drops out such textual labels. Here the textual label ``\textit{country of origin}'' for the second instance is droppoed out. Bottom: LPD directly concatenates the textual label and the context sentence. The textual label serves as a prompt to guide BERT to derive a better class prototype. Note that for simplicity we use the relation names here, while in our implementation we use relation descriptions, which are lengthier and more complex.}
		\label{fig:first-illustration}
	\end{figure}

 While the BERT-based prototypical networks \citep{baldini-soares-etal-2019-matching, peng-etal-2020-learning} have shown impressive performance on FSRE, the class prototypes are only constructed through the average representation of support instances of each class, neglecting the textual labels that may provide additional useful 
 information. Therefore, recent efforts try to modify the prototypical network such that it can use the label information as well.  \citet{DBLP:conf/cikm/YangZDHHC20} insert both entity type information and relation descriptions to the model. \citet{dong-etal-2021-mapre} use a relation encoder to generate relation representation besides the sentence encoder. \citet{han-etal-2021-exploring} propose a hybrid prototypical network that can generate hybrid prototypes from context sentences and relation descriptions. Nonetheless, these methods largely assume that every support instance is provided with a corresponding textual label in the support set during both learning and prediction. We argue that injecting textual labels to all support instances may render the training task unchallenging, because the model can largely rely on the textual labels during training, and thus results in poor performance during testing when faced with unseen relations and textual labels. Ideally, textual labels should be treated as additional source of information, such that the model can work with or without the textual labels, as shown in the top part in Figure \ref{fig:first-illustration}.

In this work, we propose a novel approach called \textit{Label Prompt Dropout} (LPD). We directly concatenate the textual label and the context sentence, and feed them together to the Transformer encoder \citep{NIPS2017_3f5ee243}. The textual label serves as a {\em label prompt}\footnote{\color{black}In this work, we use the terms {\em label prompt}, {\em relation description}, and {\em textual label} interchangeably. However, our method differs from the conventional prompt-based model in which a verbalizer \cite{schick-schutze-2021-exploiting} is needed. We use relation description to construct a natural language sentence for each instance to better make use of implicit knowledge acquired by language models during pre-training. This goal is similar to that of the conventional prompt-based method. This is why we call call our method {\em label prompt dropout}.} to guide and regularize the Transformer encoder to output a label-aware relation representation through self-attention. During training, we randomly drop out the prompt tokens to create a more challenging scenario, such that the model has to learn to work with and without the relation descriptions. Experiments show our approach achieves significant improvement on two standard FSRE datasets. Extensive ablation studies are conducted to demonstrate the effectiveness of our approach.
Furthermore, we highlight a potential issue with the evaluation setup of previous research efforts, in which the pre-training data contains relation types that actually overlap with those in the test set.
We argue that this may not be a desirable setup for few-shot learning, and show that the performance gain of existing efforts may be partly due to this ``knowledge leakage'' issue.
We propose to filter out all the overlapping relation types in the pre-training data and conduct more rigorous few-shot evaluation. In summary, we make the following contributions:
\begin{itemize}
		\item We present LPD, a novel label prompt dropout approach that makes better use of the textual labels in FSRE. This simple design has significantly outperformed previous attempts that fuse the textual label and the context sentence {\color{black}using} complex network structures.
		\item We identify the limitation of the previous experimental setup in the literature and propose a stricter setup for evaluation in FSRE. For both setups, we show strong improvements over the previous state of the art.
	\end{itemize}

\begin{figure*}[ht]
    \centering
    \includegraphics[width=1.0\textwidth, clip=true, trim = 0mm 40mm 0mm 43mm]{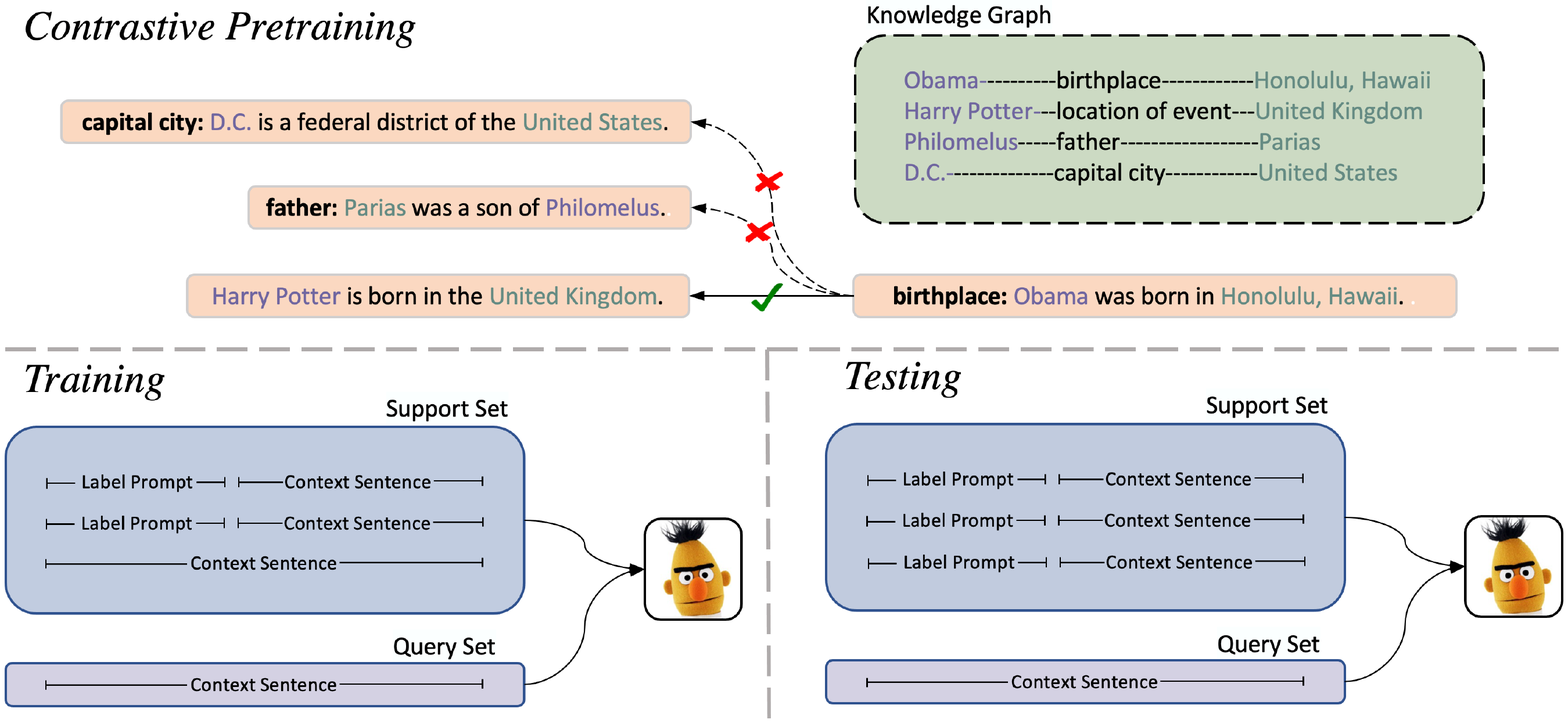}
    \caption{The framework of LPD. We prepend label prompt at the front of context sentences, and dropout the label prompt with probability \(\alpha\) (\(\alpha\textsubscript{pre-train}\), \(\alpha\textsubscript{train}\), \(\alpha\textsubscript{test}\) for the pre-training, training, and testing stage, respectively). Top: we follow \citet{peng-etal-2020-empirical} to use a knowledge graph to distantly annotate the pre-training corpus.  Bottom Left: during training, label prompt in the support set is randomly dropped out, while there is no label prompt for the query instance. Bottom right: during testing,\(\alpha\textsubscript{test}\) is set to zero, meaning that all support instances are equipped with label prompts. }
    \label{fig:figure_2}
    \vspace{-1ex}%
\end{figure*}

\section{Related Work}

\subsection{Few-Shot Relation Extraction}
Few-shot relation extraction (FSRE) aims to train a model that can classify instances into novel relations with only a handful of training examples. \citet{han-etal-2018-fewrel} are the first to introduce a large scale benchmark for FSRE, in which they evaluate a model in $N$-way-$K$-shot settings.
\citet{DBLP:conf/aaai/GaoH0S19} propose a hybrid attention-based prototypical network to handle the diversity and noise problem of text data. 
\citet{DBLP:conf/icml/QuGXT20} model the relationship between different relations via Bayesian meta-learning on relation graphs. 
\citet{han-etal-2021-exploring} apply an adaptive focal loss and hybrid networks to model the different difficulties of different relations.

Another line of work focuses on further training pre-trained language models (PLMs) on the task of relation extraction (RE). Based on the hypothesis that sentences with the same entity pairs are likely to express the same relation, \citet{baldini-soares-etal-2019-matching} collect a large-scale pre-training dataset and propose a ``matching the blanks'' pre-training paradigm. \citet{peng-etal-2020-learning} present an entity-masked contrastive pre-training framework for relation extraction.
\citet{dong-etal-2021-mapre} introduce a semantic mapping approach to include relation descriptions in the pre-training phase. Inspired by these works, we propose a contrastive pre-training with label prompt dropout approach to use relation descriptions during pre-training while creating a more difficult setup by dropping out the relation descriptions.

\subsection{Prompt-Based Fine-Tuning}
Prompt-based models have shown promising performance in few-shot and zero-shot learning in many recent studies \citep{brown2020language, schick-schutze-2021-exploiting, shin-etal-2020-autoprompt}. Models in this line of research try to align the downstream fine-tuning task with the pre-training masked language modeling objective \citep{devlin-etal-2019-bert} to better use the pre-trained language model's latent knowledge. \citet{han2021ptr} use prompt tuning with rules to perform relation classification. \citet{liu2022pretrain} introduce ``Multi-Choice Matching Networks'' that construct prompts by concatenating multiple relation descriptions. 

However, unlike many other tasks in NLP where the label semantics are straightforward, such as ``\textit{positive/negative}'' in binary sentiment analysis, the relation types in relation extraction can be quite complex, often requiring lengthy sentences as their descriptions. For example, relation P2094 in FewRel is described to be ``\textit{official classification by a regulating body under which the subject (events, teams, participants, or equipment) qualifies for inclusion}''. Prompt-based models struggle in this case because they require the template to be fixed (e.g., the number of \texttt{[MASK]} tokens in the prompt template has to be fixed). Previous approaches had to rely on manually designed prompt templates and use relation names instead of relation descriptions. To tackle this problem, we propose to directly use the entire relation description as the prompt without any mask tokens. While in conventional prompt-based models, prompts are used to create natural descriptions such that the model can perform better prediction at the \texttt{[MASK]} positions, the label prompt used in this work uses natural descriptions to help regularize the model to output a better class representation.

\section{Task Definition}

For an FSRE task, each instance \((x, e, y)\) is composed of a context sentence \(x=\{x_1, x_2, x_3, ..., x_m\}\), where \(x_i\) stands for the input token of position \(i\); entity positions \(e = \{e_{head}, e_{tail}\}\), where \(e_{head}\) refers to the head entity span and \(e_{tail}\) refers to the tail entity span; and a label \(y = \{y_{text}, y_{num}\} \), where \(y_{text}\) is the textual label and \(y_{num}\) is the numerical label.

Let \(\mathcal{E}_{train}, \mathcal{E}_{val}, \mathcal{E}_{test}\) be the training, validation, and test dataset with mutually exclusive label sets. Under the meta-learning paradigm, each dataset consists of multiple episodes, each with a support set \(\mathcal{S}\) and query set \(\mathcal{Q}\). For $N$-way-$K$-shot learning, the support set \(\mathcal{S} = \{s_k^n; n = 1, ..., N, k = 1, ..., K\}\) contains $N$ different classes. Inside each class there are $K$ different support instances. Our job is to predict the correct label \( y \in \{y^1, ..., y^N\}\) for each query instance \(q\) in the query set. In this work, we will follow the continued pre-training setup \citep{peng-etal-2020-learning}, so there is another dataset \(\mathcal{E}_{pretrain}\). Note that this \(\mathcal{E}_{pretrain}\) is not the dataset used for the masked language modeling but for the domain-specific pre-training in RE.

\section{Approach}

\subsection{Training with Label Prompt Dropout}

For each support instance, we directly concatenate the relation description and context sentence with a ``:'' in between. For example, the sentence ``\textit{Beijing held the 2022 winter Olympics}'' will become ``\textit{location of event: Beijing held the 2022 winter Olympics.}'' The idea is to create a natural instance where definition is given first, followed by examples. The relation description and colon serve as a label prompt to guide the Transformer encoder to output a label-aware relation representation. To prevent the model from relying entirely on the label prompt and overlooking the context sentence, the label prompt is randomly dropped out with probability \(\alpha\textsubscript{train}\).  For example, the support instance ``\textit{Decimal number was first developed in India}'' in Figure \ref{fig:first-illustration} remains in its initial form because its label prompt is dropped out.  For query instances, we directly input the sentence without any label prompt. This is becauase the query set is essentially the same as the test set, where we should not assume access to the ground truth knowledge. Subsequently, the special entity markers are used to mark the head and tail mentions \cite{zhang-etal-2019-ernie, baldini-soares-etal-2019-matching}, and we add the special classification and separation token to the front and the end of the sentences, such as ``\textit{\texttt{[CLS]} location of event: \texttt{[E1]} Beijing \texttt{[/E1]} held the \texttt{[E2]} 2022 winter Olympics \texttt{[/E2]} .}'' The parsed sentence is then fed to the Transformer encoder. We concatenate the final layer representations of the start entity markers (i.e., \texttt{[E1]} and \texttt{[E2]}), forming the relation representation of each instance:
\begin{equation}
     r = [ {\rm Encoder}(x)_{h};{\rm Encoder}(x)_{t}]
\end{equation}
where $h$ stands for the position of \texttt{[E1]}, $t$ stands for the position of \texttt{[E2]}, and $r$ is the relation representation. For $K$-way-$N$-shot learning, we average the relation representations of the \(K\) support instances within one class to obtain the class prototype. The dot product between the query instance and each class prototype is then calculated and used as the logit in the cross entropy loss:
\begin{eqnarray}
    &u^n = \frac{1}{K}\sum_{k=1}^{K} r^n_k 
\\
    &\mathcal{L\textsubscript{\textit{train}}} = -\sum_{n=1}^{N} \log \frac{\exp( r^{\top}_q u^{n})}{\sum_{n'=1}^{N}\exp(r^{\top}_q u^{n'})} 
\end{eqnarray}
where \(r^n_k\) stands for the relation representation of the $k$-th support instance in class \(n\), \(u^n\) is the class prototype for class \(n\), and \(r_q\) is the relation representation of the query instance.

\subsection{Testing with Prompt Guided Prototypes}
Similar to the standard dropout operation \citep{JMLR:v15:srivastava14a} in which neurons are randomly dropped out during training and are restored during testing, LPD does not drop out any label prompt of the support instances during testing as well. By inputting the relation description together with each support instance, we essentially obtain a prompt guided prototype for every support class.
We output the prediction by finding the closest class prototype to the query's relation representation: 
\begin{equation}
    \hat{y}_{num} =  \mathop{\arg\max}_{n} r^{\top}_q u^{n}\\
\end{equation}

\subsection{Contrastive Pre-training with Label Prompt Dropout}\label{4.1}

LPD can also be added to the domain-specific pre-training stage in relation extraction. In fact, pre-training is a crucial step for LPD, because the large dataset in the pre-training stage allows the model to fit to the LPD input format and learn how to extract useful information from the label prompts. We follow the framework proposed by \citet{peng-etal-2020-empirical} to sample positive and negative pairs used in contrastive pre-training. Given a knowledge graph (KG) $\mathcal{K}$ and two sentences with entity pairs (\(h_1\), \(t_1\)) and (\(h_2\), \(t_2\)), the two sentences will be labeled as a positive pair if $\mathcal{K}$ defines a relation \(R\) such that (\(h_1\), \(t_1\)) and (\(h_2\), \(t_2\)) belongs to that relation. Sentences that do not form a positive pair will be sampled as negative pairs. In Figure \ref{fig:figure_2}, for instance, the entity pairs (\textit{Harry Potter}, \textit{United Kingdom}) and (\textit{Obama}, \textit{Honolulu Hawaii}) form the same relation ``\textit{birthplace}'' in the KG. Thus, the two sentences containing these two entity pairs are sampled as a positive pair. Under such an approach, the data is noisily labeled, because two entities forming the relation in the KG may not express such relation in the sentence. For example, ``\textit{D.C. is a federal district of the United States}'' will be labeled as as an instance of the ``\textit{captical of}'' relation, even though such relation is not expressed in this sentence. Following \citet{baldini-soares-etal-2019-matching}, we randomly mask entity spans with the special \texttt{[BLANK]} token with probability $\rho_{blank} = 0.7$ to avoid relying on the shallow cues of entity mentions.

Each instance in the pre-training stage undergoes the same transformation as that of the support instances in the training stage. A label prompt is prepended to each sentence with dropout probability \(\alpha\textsubscript{pre-train}\), and special tokens are inserted to the sentences. Contrastive loss is used to train the model:
\begin{equation}
   \mathcal{L\textsubscript{\textit{CP}}} = - \log \frac{\exp( r_A r_B)}{ \exp( r_A r_B) + \sum_{i=1}^{N}\exp(r_A r_{B}^{i})} 
\end{equation}
where \((r_A, r_B)\) constitutes the positive pair and \((r_A, r_{B}^{i}), 1 \leq i \leq N\) represents negative pairs. Following \citet{peng-etal-2020-empirical}, the masked language modeling objective (\(\mathcal{L\textsubscript{MLM}}\)) is used to maintain the model's ability of language understanding. So the final pre-training loss becomes:
\begin{equation}
   \mathcal{L\textsubscript{\textit{pre-train}}} =  \mathcal{L\textsubscript{\textit{CP}}} + \mathcal{L\textsubscript{\textit{MLM}}}
\end{equation}

\section{Experimental Setup}
\subsection{Datasets} \label{dataset-seciton}
    \begin{table}
    \centering
    \scalebox{0.8}{
    \begin{tabular}{lll} \toprule
    Dataset & \# class & \# instance \\ \midrule
    Wikipedia & $698$\tablefootnote{\citet{peng-etal-2020-empirical} reported 744 but they actually filtered out relations with only one instance in their implementation, resulting in 698 relations only. We followed their implementation.} & $773,307$ \\
    Wikipedia (filtered) & $598$ & $331,445$ \\ \bottomrule
    \end{tabular}}
    \caption{Pre-trainig dataset statistics.}
    \label{table:pretraining_state}
    \end{table}

Following \citet{peng-etal-2020-empirical} and \citet{han-etal-2021-exploring}, we use FewRel 1.0 \cite{han-etal-2018-fewrel} and FewRel 2.0 (the domain adaption portion) \cite{gao-etal-2019-fewrel} to evaluate our model. FewRel 1.0 is a large-scale FSRE dataset sampled from Wikipedia articles and annotated by human annotators. It contains 100 relations and 700 instances for each relation. We follow the official split to use 64 relations for training, 16 for validation, and 20 for testing. In order to study the domain transferability of LPD, we also evaluate our model on FewRel 2.0, which is collected from the biomedical domain and does not contain a training set. It has 25 relations and 100 instances for each relation.

For contrastive pre-training, we use the same dataset as in \citet{peng-etal-2020-empirical}. This dataset is collected using the method introduced in Section \ref{4.1}, with Wikipedia articles as the corpus and Wikidata \citep{42240} as the knowledge graph. We notice that \citet{peng-etal-2020-empirical} had excluded all entity pairs in test sets of FewRel 1.0 from the pre-training data, but they did not exclude the relation types that appear in the train, validation and test set of FewRel 1.0 from the pre-training dataset. We argue that this may not be a desirable setup for few-shot learning due to the potential ``knowledge leakage'': models can learn the distant supervision signal in the pre-training dataset and thus learn the relation types in FewRel 1.0 during the pre-training stage, even though the pre-training dataset is not manually annotated by human. Thus, we propose a harder experimental setup by filtering out all 100 relation types present in FewRel 1.0, not just the entity pairs in the FewRel 1.0 test set, from the pre-training dataset. We will refer to the original dataset produced by \citet{peng-etal-2020-empirical} as Wikipedia, and the filtered out dataset as Wikipedia (filtered) in this paper. Table \ref{table:pretraining_state} shows statistics of the original pre-training dataset and the new dataset after filtering.

\subsection{Implementation Details}

    \begin{table}[t]
		\centering
		\scalebox{0.7}{
			\begin{tabular}{ccc}
				\toprule
				Stage & Parameter & Value\\
				\midrule
				\multirow{7}*{Pre-training}
				& Transformer encoder & bert-base-uncased \\
				& hidden size& $768$\\
				& max length & $64$ \\
				& learning rate & $3e-5$  \\
				& batch size& $2048$\\
				& epoch& $20$\\
				& $\alpha\textsubscript{pre-train}$ & $0.6$\\
				\midrule
				\multirow{7}*{Training}
				& Transformer encoder & bert-base-uncased \\
				& hidden size& $768$\\
				& max length & $128$ \\
				& learning rate & $2e-5$  \\
				& batch size& $4$\\
				& max iteration& $10,000$\\
				& $\alpha\textsubscript{train}$ & $0.4$ / $0.8$ \\
				\bottomrule
			\end{tabular}
		}
		\caption{\label{hyper-parameters}Hyperparameters of LPD.}
	\end{table}

We pre-train our model on top of BERT-base from the Huggingface Transformer library\footnote{\url{https://github.com/huggingface/transformers}} using 2 RTX 6000. The entire pre-training process took around 6 hours on the original Wikipedia dataset and 3 hours on the filtered one. It takes 5 hours to fine-tune our model on FewRel 1.0 with a single RTX 6000. Table \ref{hyper-parameters} shows the detailed hyperparameters. The same set of hyperparameters, except \(\alpha\textsubscript{train}\), are used for both FewRel 1.0 and FewRel 2.0. We set \(\alpha\textsubscript{train}\) to 0.4 for FewRel 1.0, and 0.8 for FewRel 2.0, which we tuned based on the model's accuracy on the validation sets.

\begin{table*}[t]
		\centering
		\scalebox{0.6}{
			\begin{tabular}{lcccccccccc}
				\toprule
                \multirow{2}[3]{*}{Model}
                & \multirow{1}{*}[-0.5\dimexpr \aboverulesep + \belowrulesep + \cmidrulewidth]{Pre-train}
                &\multirow{1}{*}[-0.5\dimexpr \aboverulesep + \belowrulesep + \cmidrulewidth]{Use LPD}
                &\multicolumn{2}{c}{5-way-1-shot}&\multicolumn{2}{c}{5-way-5-shot}&\multicolumn{2}{c}{10-way-1-shot}&\multicolumn{2}{c}{10-way-5-shot}\\
                \cmidrule(lr){4-5} \cmidrule(lr){6-7}\cmidrule(lr){8-9} \cmidrule(lr){10-11}
                & \multirow{1}{*}[0.5\dimexpr \aboverulesep + \belowrulesep + \cmidrulewidth]{Dataset} &\multirow{1}{*}[0.5\dimexpr \aboverulesep + \belowrulesep + \cmidrulewidth]{in Pre-train}& val & test& val & test& val & test& val & test\\
				\midrule
				 BERT-PAIR $^\clubsuit$  & --- & No &85.66 & 88.32 & 89.48 & 93.22 &76.84 & 80.63&81.76 & 87.02 \\
				Proto-BERT $^\varheart $  & --- & No&84.92 &89.13 & 89.03& 94.38 &76.22 & 82.77  &83.55 & 90.05 \\
				REGRAB    & --- & No&87.95 & 90.30 & 92.54 & 94.25 &80.26 &84.09&86.72 & 89.93 \\
				HCRP    & --- & No& {90.90} & 93.76   & {93.22} &
				{95.66}  & {84.11} & {89.95} & {87.79} & {92.10} \\
				LPD   & --- & No & 88.84 \(\pm\) 0.5 & {93.79} \(\pm\) 0.4 &  90.65 \(\pm\) 0.6 &  95.07 \(\pm\) 0.4 & 	79.61 \(\pm\) 0.9 & 89.39 \(\pm\) 0.5 & 82.15 \(\pm\) 1.1 & 91.08 \(\pm\) 0.6\\ 
				\midrule
				\midrule
				MTB    & Wikipedia & No&--- & 91.10 & --- & 95.40 &---  & 84.30&---  & 91.80 \\
				CP  & Wikipedia &  No &---  & 95.10 & --- & 97.10 &---  & 91.20&--- &  94.70 \\
				MapRE  & Wikipedia  & No & ---  & 95.73 & ---  & 97.84& --- & 93.18 & --- & 95.64 \\
				HCRP  & Wikipedia  & No & 94.10 & 96.42 & 96.05 & 97.96& 89.13  &  93.97 & 93.10 &  96.46 \\
				LPD    & Wikipedia & No & 95.00 \(\pm\) 0.2 & 96.29 \(\pm\) 0.0 &  96.88 \(\pm\) 0.2  &  97.34 \(\pm\) 0.1  & 92.26 \(\pm\) 0.3  & 93.33 \(\pm\) 0.3 & 94.10 \(\pm\) 0.4 & 94.63 \(\pm\) 0.2  \\
				LPD   & Wikipedia & Yes & \textbf{97.76} \(\pm\) 0.1 & \textbf{98.17} \(\pm\) 0.0 & \textbf{97.75} \(\pm\) 0.2 & \textbf{98.29} \(\pm\) 0.2 & \textbf{96.21} \(\pm\) 0.2  &  \textbf{96.66} \(\pm\) 0.0 & \textbf{96.28} \(\pm\) 0.1 &  \textbf{96.75} \(\pm\) 0.2\\
				\midrule
				CP $^\varheart $ & Wikipedia (filtered) & No & 88.29 \(\pm\) 0.2 & 90.85 \(\pm\) 0.1 &  92.77 \(\pm\) 0.6 & 95.60 \(\pm\) 0.3 & 80.50\(\pm\) 0.6 & 83.89\(\pm\) 0.9 & 88.61 \(\pm\) 0.3 & 90.61 \(\pm\) 0.3\\ 
				HCRP $^\varheart$ & Wikipedia (filtered) & No & 90.71 \(\pm\) 0.4 & 93.32 \(\pm\) 0.7 &   93.54 \(\pm\) 0.3 & 95.33 \(\pm\) 0.5 &  84.35 \(\pm\) 0.9 & 88.72 \(\pm\) 0.7 & 88.64 \(\pm\) 0.6 & 91.71 \(\pm\) 0.5 \\ 
				LPD & Wikipedia (filtered) & No & 90.89 \(\pm\) 0.1 & 94.23 \(\pm\) 0.1 &  92.90 \(\pm\) 0.3  & 95.77 \(\pm\) 0.2 & 83.17 \(\pm\) 0.3 & 89.69 \(\pm\) 0.1 & 86.43 \(\pm\) 0.7 & 91.94 \(\pm\) 0.2 \\ 
				LPD & Wikipedia (filtered) & Yes  & \textbf{93.51} \(\pm\) 0.7& \textbf{95.12} \(\pm\) 0.2 & \textbf{94.33} \(\pm\) 0.7 & \textbf{95.79} \(\pm\) 0.1 & \textbf{87.77} \(\pm\) 1.1 & \textbf{90.73} \(\pm\) 0.2 & \textbf{89.19} \(\pm\) 1.3 & \textbf{92.15} \(\pm\) 0.3 \\
				\bottomrule
			\end{tabular}
		}
		\caption{Accuracy (\%) of few-shot classification on the FewRel 1.0 validation / test set. Top: models that are directly trained on FewRel 1.0 with BERT-base. Middle: models that are pre-trained on the original Wikipedia dataset \cite{peng-etal-2020-learning}. Bottom: models that are pre-trained on the Wikipedia (filtered) dataset as discussed in section \ref{dataset-seciton}. $\clubsuit$ are from FewRel public leaderboard\footnotemark[2], and $\varheart $ are produced by our implementation. }
		\label{main-fewrel1}
	\end{table*}

\begin{table}
		\centering
		\renewcommand\tabcolsep{3.6pt}
		\scalebox{0.55}{
			\begin{tabular}{lccccc}
				\toprule
				\multirow{2}*{Model} 
				&Pre-train&5-way&5-way&10-way& 10-way\\
				&Dataset&1-shot&5-shot&1-shot& 5-shot\\
				\midrule 
				Proto-CNN & --- &35.09 &49.37  &22.98  &35.22  \\
				Proto-BERT   & --- &40.12  & 51.50  &26.45 &36.93  \\
				Proto-ADV  & --- & 42.21 &  58.71 &28.91&44.35 \\
				BERT-PAIR & --- &67.41 & 78.57&54.89&66.85 \\
				HCRP  & --- & 76.34 &83.03  & 63.77&72.94 \\
				LPD & --- & \textbf{77.82} \(\pm\) 0.4 & \textbf{86.90} \(\pm\)0.3 & \textbf{66.06} \(\pm\) 0.6 & \textbf{78.43}  \(\pm\) 0.4\\
				\hline
				CP  & Wikipedia & 79.70 & 84.90  & 68.10 & 79.80 \\
				CP $\varheart $ &Wikipedia (filtered) & 80.35 \(\pm\) 1.2 & 88.69 \(\pm\) 0.5 & 69.33 \(\pm\) 1.7 & 80.95 \(\pm\) 1.0 \\
				LPD & Wikipedia & 82.81 \(\pm\) 0.5 & 88.98 \(\pm\) 1.4 & 70.51 \(\pm\) 1.5 & 78.76  \(\pm\) 1.6 \\
				LPD & Wikipedia (filtered) & \textbf{83.41} \(\pm\) 0.5 & \textbf{90.00} \(\pm\) 0.3 & \textbf{73.28} \(\pm\)  0.8 & \textbf{81.80 }  \(\pm\) 0.9\\
				\bottomrule
			\end{tabular}
		}
		\caption{Accuracy (\%) of few-shot classification on the FewRel 2.0 test set. $ \varheart $ are produced by our implementation. }
		\label{main-fewrel2}
		
	\end{table}

\subsection{Evaluation}

We evaluate our model by randomly sampling 10,000 episodes from the $N$-way-$K$-shot support set and a query instance. We follow previous works \citep{han-etal-2018-fewrel, gao-etal-2019-fewrel} to choose $N$ to be 5 and 10, and $K$ to be 1 and 5. For the main comparison (i.e., Table \ref{main-fewrel1} and Table \ref{main-fewrel2}), we report the average accuracy together with the standard deviation of 3 runs using different random seeds. We report the accuracy of 1 run for all ablation studies. To obtain the test set accuracy, we submit our predictions to the FewRel leaderboard\footnote{\url{https://competitions.codalab.org/competitions/27980}}.

\section{Results}

\begin{table*}[ht]
		\centering
		\renewcommand\tabcolsep{3.6pt}
		\scalebox{0.8}{
			\begin{tabular}{lccccccccccccc}
				\toprule
				 \(\alpha\textsubscript{pre-train}\)        & 0.0  & 0.001  & 0.02 & 0.07 & 0.1   & 0.2   & 0.4   & 0.6   & 0.8   & 0.9 & 0.99 & 0.999 & 1.0 \\
				\midrule
				Accuracy  & 47.78 & 63.15 & 75.91 & 75.84 & 75.77 & 76.02 & 76.28 & 75.21 & 74.98 & 75.39 & 74.99& 74.02 & 73.46 \\

				\bottomrule
			\end{tabular}
		}
		\caption{10-way-1-shot accuracy on FewRel 1.0 validation set after pre-training with different \(\alpha\textsubscript{pre-train}\). We directly evaluate the pre-trained model on the validation set without any training process.}
		\label{pre-train-drop}
		\vspace{-1ex}%
	\end{table*}

\subsection{Comparison with Baselines} \label{sec:comparison}

We compare our model with the following baseline methods: 1) \textbf{Proto-BERT} \citep{DBLP:conf/nips/SnellSZ17} is a prototypical network with BERT-base \citep{devlin-etal-2019-bert} serving as the backbone. Note that our proposed method will be reduced to Proto-BERT if we discard pre-training and set \(\alpha\textsubscript{train}\) and \(\alpha\textsubscript{test}\) to 1.0. 2) \textbf{BERT-PAIR} \cite{gao-etal-2019-fewrel} is a method that measures similarity of a sentence pair. 3) \textbf{REGRAB} \cite{DBLP:conf/icml/QuGXT20} is a label-aware method that models the relationship between different relation types via a Bayesian network. 3) \textbf{MTB} \cite{baldini-soares-etal-2019-matching} is a model pre-trained with their proposed matching the blank objective based on BERT. Note that we report the results produced by \citet{peng-etal-2020-learning} for a fair comparison with all BERT-base based models because \citet{baldini-soares-etal-2019-matching}'s original work is based on BERT-large. 4) \textbf{CP} \cite{peng-etal-2020-learning} pre-trains Proto-BERT using a contrastive pre-training approach that regards sentences with the same relations as positive pairs and other instances in the same batch as the negative pairs. 5) \textbf{MapRE} \citep{dong-etal-2021-mapre} extends CP with a relation encoder to consider relation type information. 6) \textbf{HCRP} \cite{han-etal-2021-exploring} equips Proto-BERT with a hybrid attention module and a task adaptive focal loss.

\paragraph{Wikipedia (filtered) is more challenging.} Comparing models that use the original Wikipedia to pre-train (the middle part of Table \ref{main-fewrel1}) and those that use the filtered version of Wikipedia (the bottom part of Table \ref{main-fewrel1}), we observe that the accuracy drops drastically across all models,  substantiating our speculation that the performance gain of existing pre-training efforts is partly due to the ``knowledge leakage'' between the pre-training dataset and the FewRel 1.0 dataset. Therefore, we call for attention on this issue.
We suggest that the community should focus more on the new setup that we propose to perform more rigorous evaluations on the FSRE task in the future.

\paragraph{LPD makes better use of textual labels.} When LPD is also used in the pre-training stage (see the last row of the bottom two blocks of Table \ref{main-fewrel1}), we find that our model significantly outperforms the previous methods, no matter if the pre-training is performed on the original Wikipedia dataset or on the filtered version. Specifically, HCRP and MapRE also use textual label in their model, but their models lead to sub-optimal performance, proving the effectiveness of dropping out textual labels during training.
Remarkably, when compared with the previous state-of-the art HCRP, LPD improves the 10-way-1-shot task by 7.08 points on the validation set and 2.69 points on the test set when the original Wikipedia dataset is used in the pre-training stage.

\paragraph{Pre-training with LPD is important.} Interestingly, we find that our model does not outperform HCRP when it is not pre-trained with LPD (in other words, when LPD is only used during training and testing), as shown in the three LPD setups with no LPD in pre-train in Table \ref{main-fewrel1}.
We hypothesize this is because our method introduces a new input format (i.e., label prompt followed by the context sentence).
As a result, it would be more beneficial for our approach to have access to abundant data in the pre-training stage, in order to learn how to acquire the relevant information within the data of such a new format.
We carry out a more detailed analysis regarding this matter in Section \ref{number-of-classes}.

\paragraph{Knowledge leakage leads to  performance drop in domain transfer.} \textcolor{black}{To evaluate LPD's ability to transfer knowledge to datasets in a different domain, we train our model on the FewRel 1.0 training set and evaluate its accuracy on the FewRel 2.0 test set. In Table \ref{main-fewrel2}, we show that LPD substantially improves the results over the previous state-of-the-art models with or without pre-training. An important finding is that when LPD is pre-trained with the unfiltered Wikipedia, its accuracy is even lower than the LPD pre-trained on the filtered counterpart, even though the former is more than twice the size of the latter (see Table \ref{table:pretraining_state}). We also run CP \cite{peng-etal-2020-learning} pre-trained with Wikipedia (filtered), and observe a similar trend to that of LPD. This again confirms our speculation that much of the previous work's performance gain on FewRel 1.0 comes from the overlapping relation types between the pre-training dataset and FewRel 1.0. When the model is pre-trained with the unfiltered Wikipedia, it overfits to the overlapping relations. While such overfitting increases the accuracy on FewRel 1.0 because the FewRel 1.0 test set also contains those overlapping relations, it fails to generalize to FewRel 2.0 where all relation types are truly novel and come from a different domain.}

\subsection{Ablation Study on the Dropout Rate}
\begin{figure}[t!] 
		\centering
		\includegraphics[width=1\linewidth,  clip=true, trim = 30mm 30mm 30mm 45mm ]{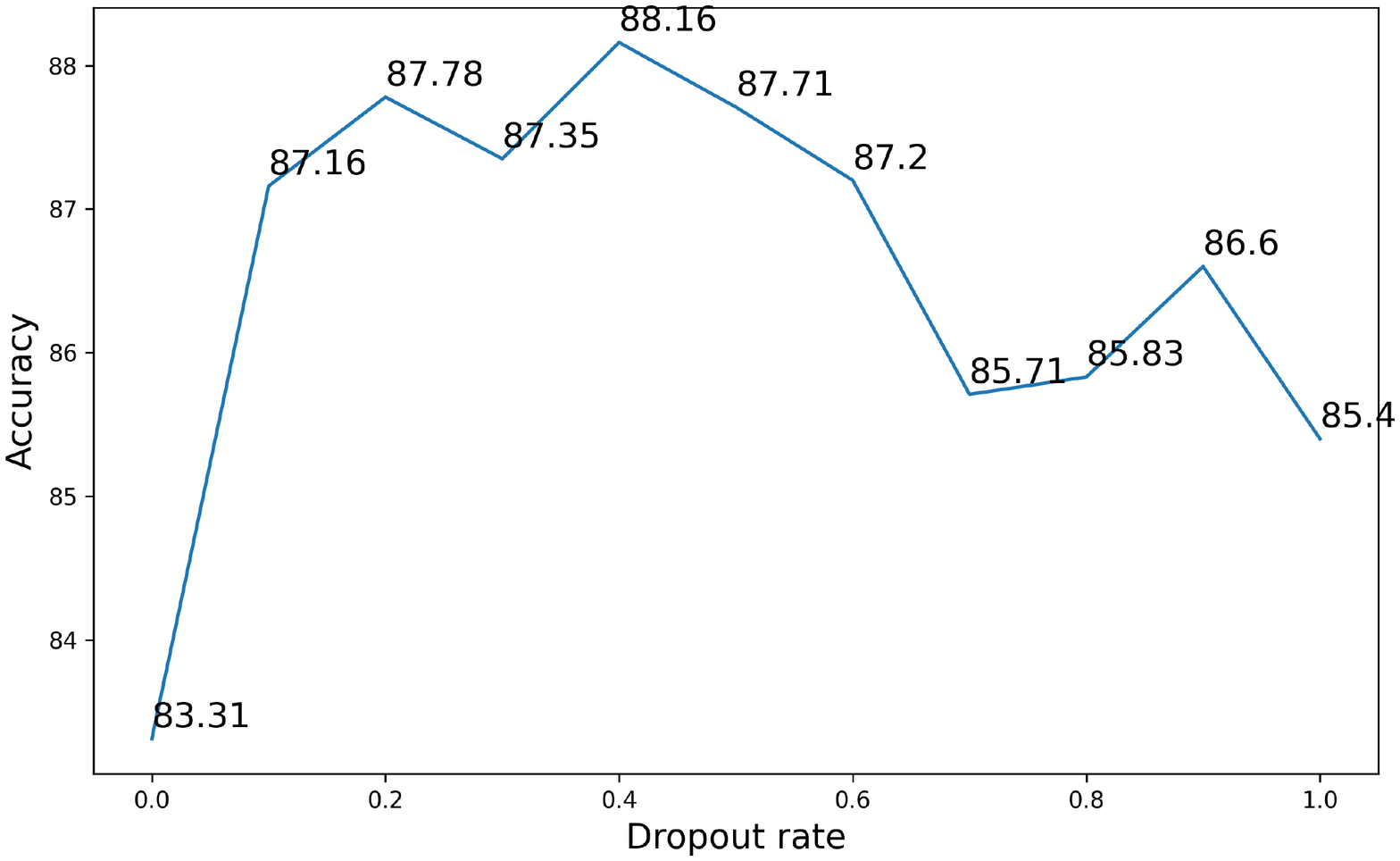}
		\caption{10-way-1-shot accuracy on FewRel 1.0 validation set after training with varying \(\alpha\textsubscript{train}\). \(\alpha\textsubscript{pre-train}\) is set to 0.6, and Wikipedia (filtered) is used.}
		\label{fig:train-alph}
	\end{figure}

The dropout rate \(\alpha\) is a crucial hyperparameter in LPD. In this section, we first study the effect of \(\alpha\textsubscript{pre-train}\) by pre-training our model on the Wikipedia (filtered) dataset, and testing it on the FewRel 1.0 validation set without any training process. As shown in Table \ref{pre-train-drop}, LPD is not very sensitive to \(\alpha\textsubscript{pre-train}\). We can only see significant drop if \(\alpha\textsubscript{pre-train}\) is below 0.1 or above 0.9. When the label prompt is always fed together with the context sentence (i.e., \(\alpha\textsubscript{pre-train}\) = 0.0), the model accuracy is rather sub-optimal. This substantiates our claim that the model will be over-reliant on the textual labels if textual labels are always fed to the model together with the context sentences. Setting dropout rate to be larger than 0.0 are essentially making the learning more challenging (as compared to \(\alpha\textsubscript{train 
}\) = 0.0), which forces the model to be more robust.  When we set \(\alpha\textsubscript{pre-train}\) to 1.0, our model essentially reduces to the contrastive learning framework proposed by \citet{peng-etal-2020-empirical}, which also results in a lower accuracy, because the model does not have access to the helpful textual label information. Figure \ref{fig:train-alph} shows the accuracy under different \(\alpha\textsubscript{train}\) when we fix \(\alpha\textsubscript{pre-train}\) to be 0.6. We can observe a similar trend as that of \(\alpha\textsubscript{pre-train}\). The model maintains a rather consistent performance when \(\alpha\textsubscript{train}\) is between 0.1 and 0.6. The accuracy drops when \(\alpha\textsubscript{train}\) is close to 0 or 1.

\subsection{Ablation Study on the Number of Relation Types} \label{number-of-classes}

\begin{figure}[t!]
		\centering
		\includegraphics[width=1\linewidth,  clip=true, trim = 30mm 30mm 30mm 45mm]{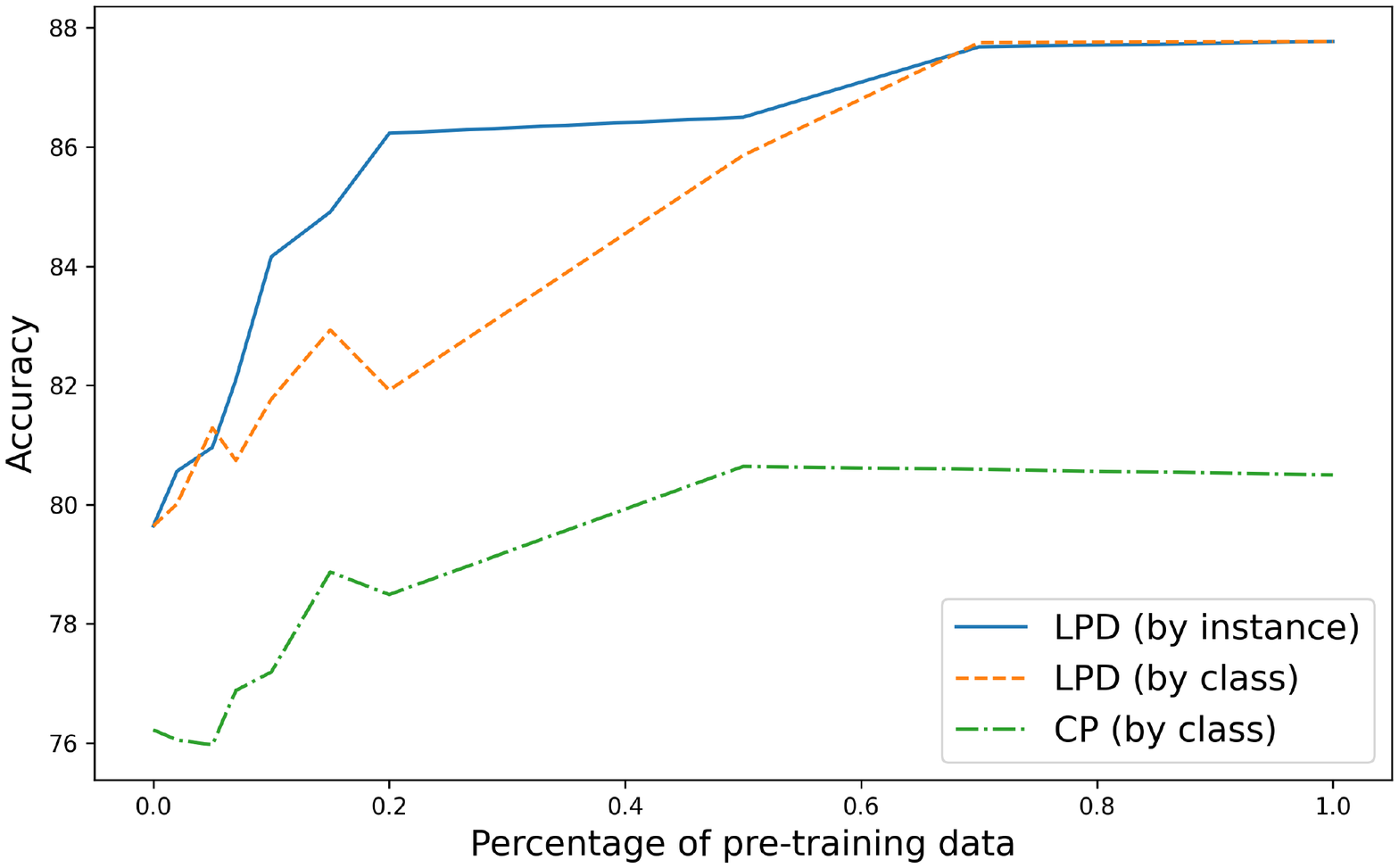}
		\caption{10-way-1-shot accuracy on FewRel 1.0 validation set with varying percentage of the the Wikipedia (filtered) used in pre-training.}
		\label{fig:class} 
	\end{figure}

As discussed in Section \ref{sec:comparison}, our method performs comparatively to HCRP when it is not pre-trained with LPD, while it outperforms all baselines when we include LPD in the pre-training stage. We hypothesized that this is because our method introduces 
a new input format (i.e., label prompt followed by
the context sentence). The new format requires our model to use the label prompt to guide the Transformer self-attention to output a better representation while still outputting high quality representations when the label prompt is dropped out. Thus, the amount of relation types in FewRel 1.0 may not be sufficient for the model to make full use of LPD. To validate this hypothesis, we only use part of the Wikipedia (filtered) dataset to pre-train the models, as shown in Figure \ref{fig:class}. \textit{LPD (by instance)} shows LPD pre-trained with $X$\% of the instances in the pre-training dataset for each relation type. We also considered another setup where we use $X$\% relation types, shown by \textit{LPD (by class)} and \textit{CP (by class)}. Comparing the \textit{by instance} and \textit{by class} setup for LPD, we find that the model is able to achieve a high accuracy with very small portion of the data when we keep the number of relation types fixed (by instance), while the performance remains sub-optimal if we only use a small set of the relation types (by class). This shows that instead of the absolute amount of pre-training instances, large number of relation types will be more beneficial to our model, confirming our hypothesis above. Comparing the by-class setup between CP and LPD, we find that the accuracy of CP saturates when 50\% relation types are used, while LPD only saturates when 70\% relation types are used for pre-training. This shows that LPD indeed needs more relation types to be properly trained, while it also has the better potential to benefit from a large amount of relation types. The accuracy increase brought by the pre-training for LPD is 7.86\%, while for CP it is 4.37\%.

\subsection{Analogy to Dropout} 

\begin{table}[t!]
		\centering
		\renewcommand\tabcolsep{3.6pt}
		\scalebox{0.7}{
			\begin{tabular}{lccc}
				\toprule
				\multirow{2}*{Model}  &\multirow{2}*{No.} &10-way-1-shot\\ &&accuracy\\
				\midrule
				LPD w/o per-taining (\(\alpha\textsubscript{train} = 0.4, \alpha\textsubscript{test} = 0.0\) )       & 1 & 79.61 \\
				Proto-BERT (\(\alpha\textsubscript{train} = 1.0, \alpha\textsubscript{test} = 1.0\) )    & 2 & 76.22 \\
				\(\alpha\textsubscript{train} = 1.0, \alpha\textsubscript{test} = 0.0\)  & 3  &  75.98 \\
				\(\alpha\textsubscript{train} = 0.4, \alpha\textsubscript{test} = 1.0\) & 4   &  75.59 \\
	            \(\alpha\textsubscript{train} = 0.4, \alpha\textsubscript{test} = 0.4\) & 5   &  77.44 \\
				corrupt description (50\%)  & 6   & 75.32 \\
				shuffle description      & 7 &  74.84 \\
				\bottomrule
			\end{tabular}
		}
		\caption{Ablation study on FewRel 1.0 validation set without pre-training.}
		\label{ablation-3}
		\vspace{-2ex}%
	\end{table}

In this section, we show that LPD shares similar properties to that of the dropout method used in neural networks \citep{JMLR:v15:srivastava14a}. In Table \ref{ablation-3}, model 3 essentially reduces to Proto-BERT during training, but it has access to the label prompt during testing. Model 3's lower performance as compared with LPD shows that it fails to make good use of the label prompt, as the model is not specifically trained for such input during learning. In model 4, we equipped the model with LPD during training, with \(\alpha\textsubscript{train} \) set to 0.4, but did not insert any label prompts during prediction. Again, the performance reduces to the same level as that of Proto-BERT. This means that LPD during training does not result in a better model to encode the relation representation if there is no label prompt prepended to the context sentence during inference. In addition,  model 5 shows a drop in accuracy from 79.61 to 77.44, demonstrating that it is crucial to set \(\alpha\textsubscript{test} \) to 0.0 to enable the model to extract useful information from the label prompt for all support instances during prediction. Using the same dropout rate during both training and testing will lead to a sub-optimal performance. These five different setups show that, interestingly, our proposed LPD approach really shares similar properties to that of the standard dropout method, which drops out a subset of but not all of the neurons during learning, and use all the neurons during prediction. 
\subsection{Analogy to Prompt}

To demonstrate that the label prompt really serves as the prompt, we first try to corrupt the information contained inside the label prompt, shown in Table \ref{ablation-3} as model 6 and model 7. In model 6, we corrupt the label prompt by randomly deleting 50\% of the tokens of each relation description.\footnote{\color{black}The deleted tokens are fixed once decided for each relation description during the entire training process.} In model 7, instead of assigning the correct relation description to the context sentence, we shuffle the descriptions and randomly assign them to the sentences.\footnote{\color{black}The mapping between the correct relation description and the randomly assigned relation description is also fixed once decided throughout the entire learning procedure.} We observed severe decline in performance in both cases, showing that adding informative and correct label prompts are crucial for guiding the model to outoput a better relation representation. This property is similar to the prompt-based model, where the prompt design should be coherent to the task and guide the model to make prediction at the \texttt{[MASK]} spans \citep{brown2020language, liu2021pre}.

\begin{figure}[t]
		\flushleft
		\includegraphics[width=\linewidth,  clip=true, trim = 0mm 40mm 0mm 40mm ]{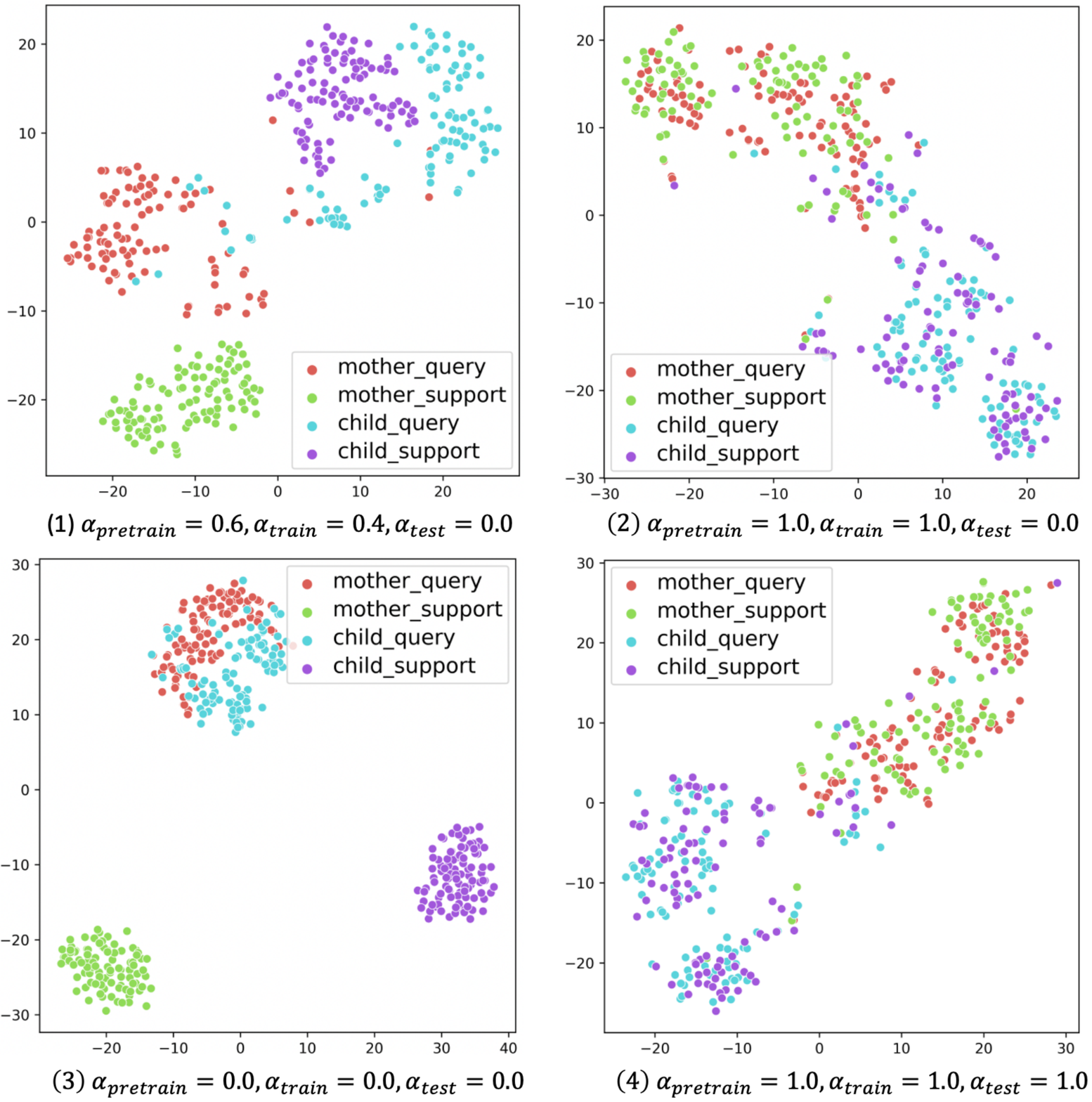}
		\caption{The t-sne plot of relation representations of two similar relations \textit{child} and \textit{mother}. \textcolor{black}{Best viewed in colour}. (1): LPD. (2): only use textual labels during testing. (3): use textual labels without dropout. (4): CP \cite{peng-etal-2020-learning}.}
		\label{metric-space} 
		\vspace{-2ex}%
	\end{figure}

To provide a qualitative examination of the effectiveness of LPD, we pre-train LPD on Wikipedia (filtered), train it on the FewRel 1.0 training set, and visualize the relation representations of two similar relations \textit{child} and \textit{mother} from the FewRel 1.0 validation set using t-sne \citep{JMLR:v9:vandermaaten08a}.
As shown in Figure \ref{metric-space}, our method is able to yield a large margin between the support instances of these two relations while maintaining a reasonable distance between query instances of these two relations. On the other hand, using relation descriptions without dropout will render the model relying too much on the relation descriptions, resulting in poor separation between query instances of the two relations, even though it achieves perfect separation for support instances. 
If we only use label prompts during testing, or simply discard all the label prompts such that the model is reduced to CP \cite{peng-etal-2020-learning}, we can observe that some of the support instances of the two relations still fall in close proximity.
We argue that the relation description really serves as a prompt to guide and regularize the support instance representations within the same class, while the model still retains the ability to work without the label prompt in the query set. Label prompts are able to promote the generation of discriminative class representations, which will benefit the FSRE task.

\section{Conclusion}
This paper proposes a novel label prompt dropout approach that directly concatenates the label prompt with the context sentence for few-shot relation extraction. The label prompt is randomly dropped out during pre-training and training to create a more challenging learning setup, leading to better use of the relation descriptions. In the experiments, we discover a ``knowledge leakage'' issue in the previous works' experimental setup. We propose a stricter setup for more rigorous evaluations in FSRE by filtering out all overlapping relations. Our method has demonstrated significant improvements on both evaluation settings. Ablation studies show that LPD shares some similar and interesting properties to the \textcolor{black}{neural dropout operation and prompt based methods. One possible direction of future work is to generalize this idea to other text classification tasks such as intent classification \cite{larson-etal-2019-evaluation-custom}}.


\section*{Limitations}
\textcolor{black}{
There are several limitations of this work.
First, LPD only works under the $N$-way-$K$-shot setup, because it requires a support set in which the textual labels are given to construct the label prompts.
Second, its effectiveness is only examined on the task of few-shot relation extraction, while whether this method is able to generalize to other text classification tasks, such as intent classification and news classification, is not yet explored in this paper.
}
Third, whether the model has the ability to perform non-of-the-above detection \cite{gao-etal-2019-fewrel}, where a query instance may not belong to any class in the support set, is not investigated in this work.

\section*{Acknowledgements}

We would like to thank the anonymous reviewers, our meta-reviewer, and senior area chairs for their constructive comments and support on our work.
We would also like to thank Vanessa Tan for her help with this work.
This research/project is supported by the National Research Foundation Singapore and DSO National Laboratories under the AI Singapore Program (AISG Award No: AISG2-RP-2020-016).


\bibliography{anthology,custom}
\bibliographystyle{acl_natbib}

\end{document}